\documentclass[sigconf]{acmart}

\usepackage{subcaption}
\usepackage{color}
\usepackage{xspace} %

\AtBeginDocument{%
  \providecommand\BibTeX{{%
    \normalfont B\kern-0.5em{\scshape i\kern-0.25em b}\kern-0.8em\TeX}}}

\copyrightyear{2022}
\acmYear{2022}
\setcopyright{rightsretained}
\acmConference[SA '22 Technical Communications]{SIGGRAPH Asia 2022 Technical Communications}{December 6--9, 2022}{Daegu, Republic of Korea}
\acmBooktitle{SIGGRAPH Asia 2022 Technical Communications (SA '22 Technical Communications), December 6--9, 2022, Daegu, Republic of Korea}\acmDOI{10.1145/3550340.3564230}
\acmISBN{978-1-4503-9465-9/22/12}

\acmSubmissionID{tcom\_130}

\begin{document}

\title{Training-Free Neural Matte Extraction for Visual Effects}
\author{Sharif Elcott}
\affiliation{%
  \institution{Google}
  \country{Japan}
}
\authornote{equal contribution.}
\email{selcott@google.com}

\author{J.P.~Lewis}
\affiliation{%
  \institution{Google Research}
  \country{USA}
}
\authornotemark[1]
\email{jplewis@google.com}

\author{Nori Kanazawa}
\affiliation{%
  \institution{Google Research}
  \country{USA}
}
\email{kanazawa@google.com}

\author{Christoph Bregler}
\affiliation{%
  \institution{Google Research}
  \country{USA}
}
\email{bregler@google.com}

\renewcommand{\shortauthors}{Elcott and Lewis, et al.}

\begin{abstract}
Alpha matting is widely used in video conferencing as well as in movies, television, and social media sites. Deep learning approaches to the matte extraction problem are well suited to video conferencing due to the consistent subject matter (front-facing humans), however training-based approaches are somewhat pointless for entertainment videos where varied subjects (spaceships, monsters, etc.) may appear only a few times in a single movie -- if a method of creating ground truth for training exists, just use that method to produce the desired mattes. We introduce a \emph{training-free} high quality neural matte extraction approach that specifically targets the assumptions of visual effects production. Our approach is based on the deep image prior, which optimizes a deep neural network to fit a single image, thereby providing a deep encoding of the particular image. We make use of the representations in the penultimate layer to interpolate coarse and incomplete "trimap" constraints. Videos processed with this approach are temporally consistent. The algorithm is both very simple and surprisingly effective. 

\end{abstract}



\begin{teaserfigure}
 \centering
  \begin{subfigure}{0.19\linewidth}
  \includegraphics[width=\textwidth]{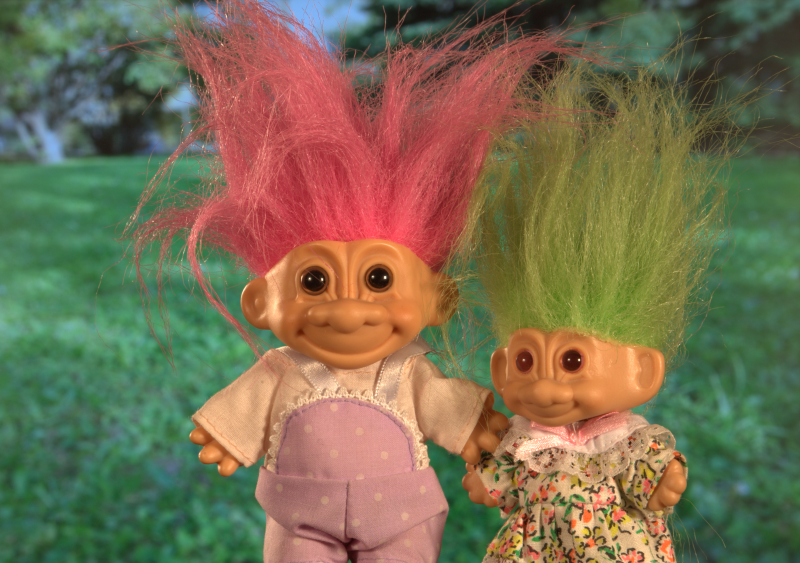}
  \caption{}   %
  \end{subfigure}
  \hfill
  \begin{subfigure}{0.19\linewidth}
  \includegraphics[width=\textwidth]{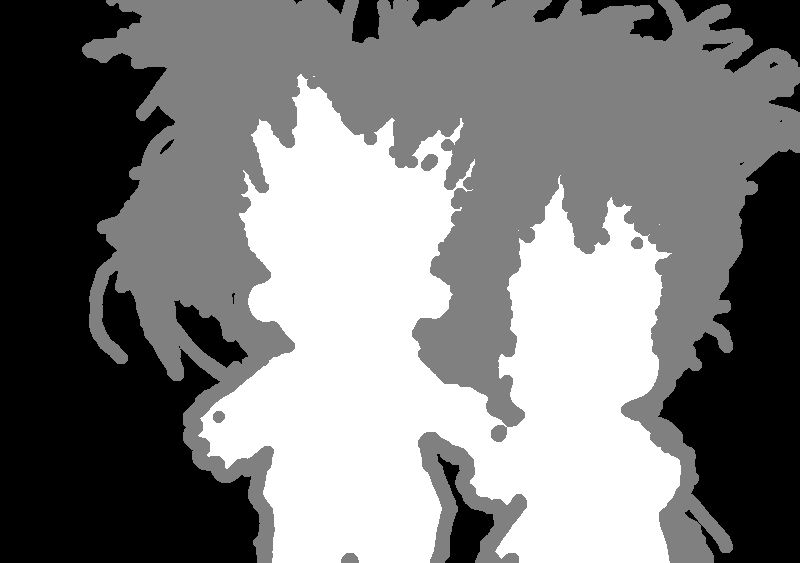}
  \caption{} 
  \end{subfigure}
  \hfill
  \begin{subfigure}{0.19\linewidth}
  \includegraphics[width=\textwidth]{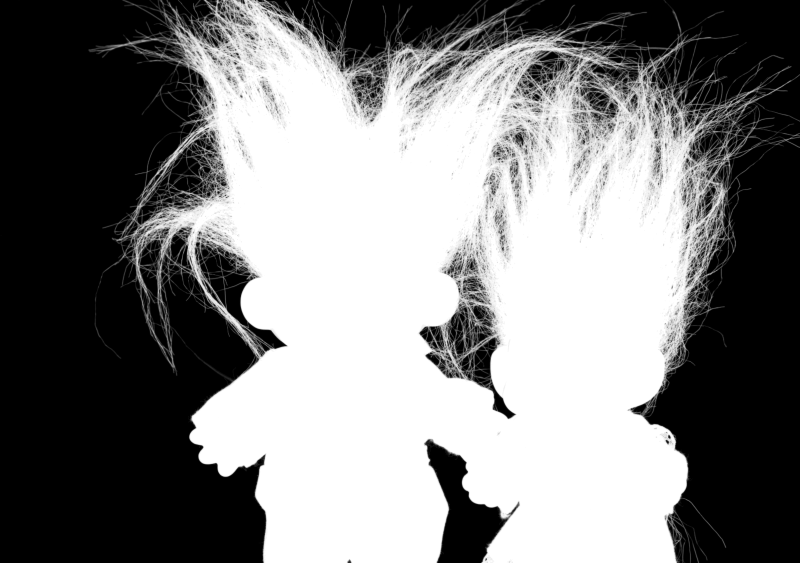}
  \caption{} 
  \end{subfigure}
  \hfill
  \begin{subfigure}{0.19\linewidth}
  \includegraphics[width=\textwidth]{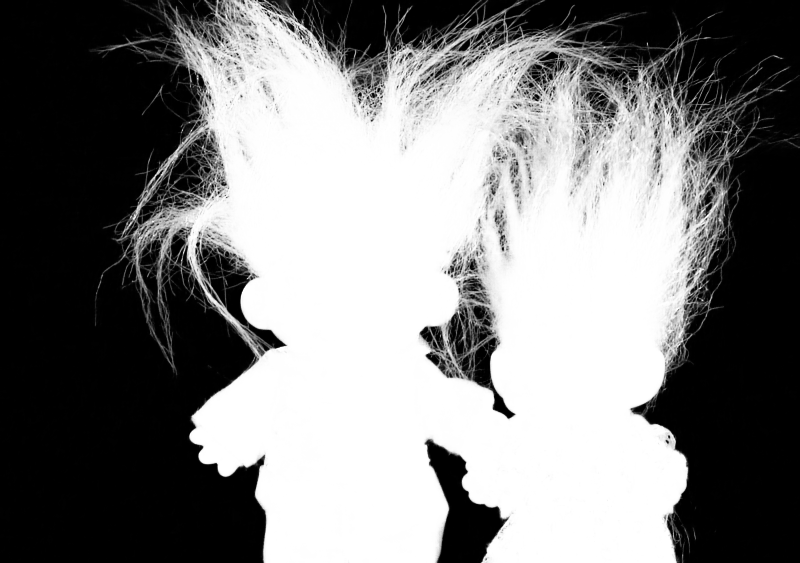}
  \caption{} 
  \end{subfigure}
  \hfill
  \begin{subfigure}{0.19\linewidth}
  \includegraphics[width=\textwidth]{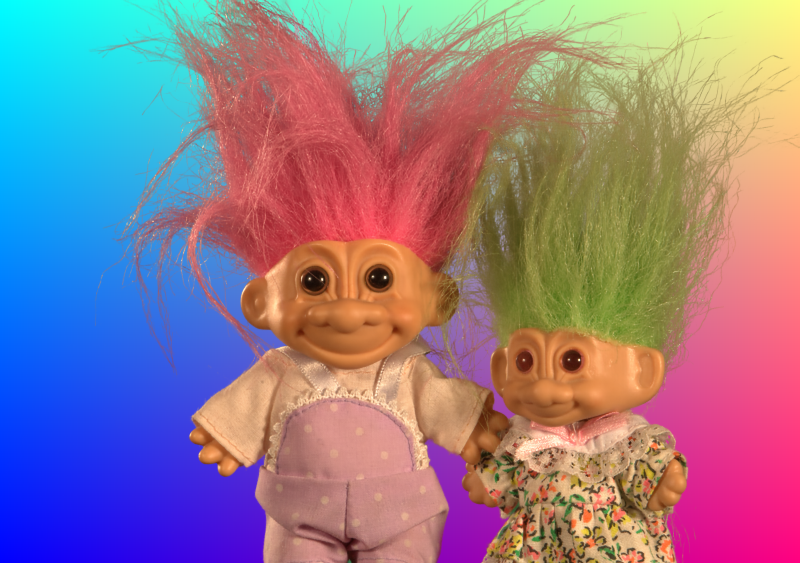}
  \caption{} 
  \end{subfigure}
  \caption{Given an image (a) and a crude "trimap" segmentation (b), we estimate a high quality alpha matte (d), allowing the foreground objects to be composited over arbitrary backgrounds (e). Image (c) shows the ground-truth alpha matte. Please enlarge to see details.}
  \label{fig:teaser}
\vspace{0.2cm} %
\end{teaserfigure}

\maketitle

\section{Introduction}

Alpha matte extraction refers to the underconstrained inverse problem of finding the unknown translucency or coverage $\alpha$ of a foreground object \cite{wang-survey-07}.
Matte extraction is widely 
used to provide alternate backgrounds for video meetings, as well as
to produce visual effects (VFX) for movies, television, and social media.
However, it is not always recognized in the research literature 
that these two applications have significantly different requirements.
This paper introduces a neural matte extraction method specifically addressed to the assumptions and requirements of VFX. 

Matting for video calls requires real-time performance and assumes a single class of subject matter, front-facing humans. This is a "train once and use many times" situation for which it is possible and advantageous to obtain training data. Video call matting often assumes a fixed camera, and may require a "clean plate" image of the room without the participant.
On the contrary, VFX production has the following assumptions and requirements:
\begin{itemize}
 \item 
   \textbf{Diverse and often rare ("one-off") subject matter}. For example the youtube video \cite{youtube} (30M views)  involves mattes of a cat, ships, and wreckage. A physical prop such as a crashed alien spaceship might only be used in a few seconds in a single movie, never to be seen again. Thus, \textbf{gathering a ground-truth training dataset for deep learning is often pointless}: if there is a method to generate the ground-truth mattes for training, just use this method to produce the desired mattes -- no need to train a model!
  \item
   Visual effects frequently involves \textbf{moving cameras} as well as extreme diversity of moving backgrounds. %
   For example, an actor might be filmed as they run or travel in a vehicle any place on earth, or on set with extraterrestrial props in the near background.
   Thus, even for the %
   major case of repeated subject matter (humans), gathering a representative training dataset is more challenging than in the case of video calls.
  \item 
    Real-time performance is not required. Instead, the \textbf{on-set filming time is to be minimized,} due to the combined cost of the actors (sometimes with 7-figure salaries) and movie crew. It is often cheaper to employ an artist to work days to "fix it in post" rather than spend a few minutes on-set (with actors and crew waiting) to address an issue. %
 \item
  \textbf{Clean-plates are not a desirable approach} for matting, and are often not possible.
With moving cameras, a motion control rig is required to obtain a clean plate. The indoor use of motion control for %
clean plates is expensive due to the previous principle (minimizing on-set time). Moreover, this approach is generally not feasible outdoors, for reasons including background movement (e.g.~plants moved by wind) and changing lighting (moving clouds blocking the sun).

\end{itemize}

Our solution, a matte extraction approach employing the deep image prior \cite{DIP-ulyanov-2018}, 
addresses these requirements. It has the following characteristics and contributions: 
\begin{itemize}
\item It is a deep neural network matte extraction method targeted towards the requirements of VFX rather than video meetings.
To our knowledge, this is the first deep neural matte extraction method
that requires neither training data nor expensive on-set image capture to support the matting process (i.e.~clean plates or greenscreens). 
\item It relies exclusively on the given image or video
along with coarse "trimaps" (Fig.~1) that are easily created during post production
using
existing semi-automatic tools. Professional software such as Nuke or After Effects is typically used, however a familiar example is the "Select Subject" tool in Photoshop \cite{photoshop-selectsubject},
followed by dilation/erosion and automated with "Actions".
\item It does not require a clean plate, thus allowing extractions from outdoor footage.
\item It does not require greenscreens, yet can produce detailed high quality mattes even when the foreground and background subject have similar colors (green hair in Fig.~1).
\end{itemize}

\newcommand{\sixwide}{0.16}
\begin{figure*}
 \centering
  \begin{subfigure}{\sixwide\linewidth}
  \includegraphics[width=\textwidth]{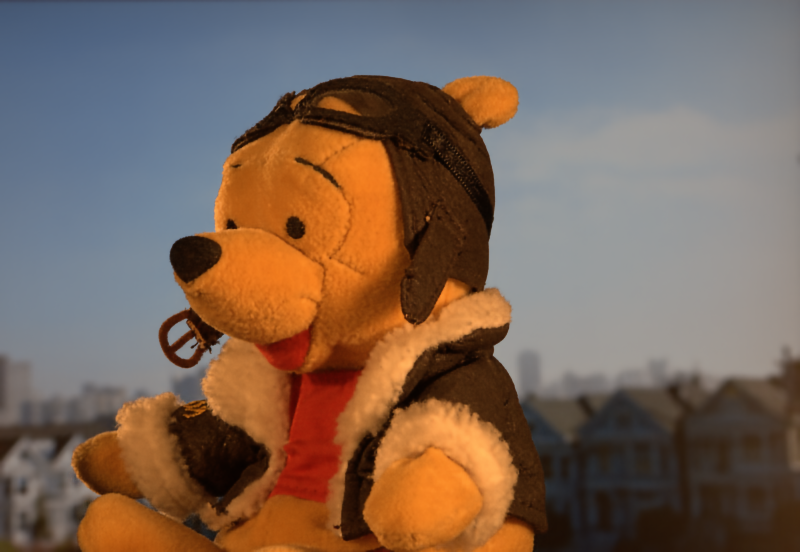}
  \end{subfigure}
  \hfill
  \begin{subfigure}{\sixwide\linewidth}
  \includegraphics[width=\textwidth]{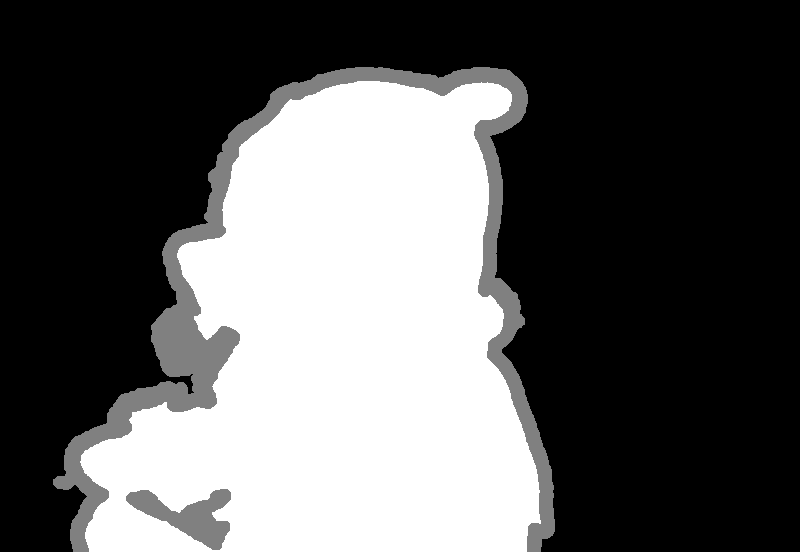}
  \end{subfigure}
  \hfill
  \begin{subfigure}{\sixwide\linewidth}
  \includegraphics[width=\textwidth]{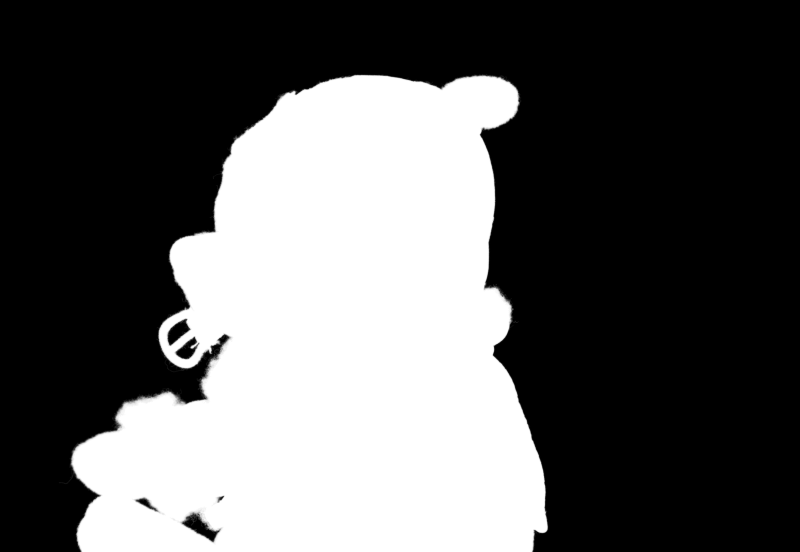}
  \end{subfigure}
  \hfill
  \begin{subfigure}{\sixwide\linewidth}
  \includegraphics[width=\textwidth]{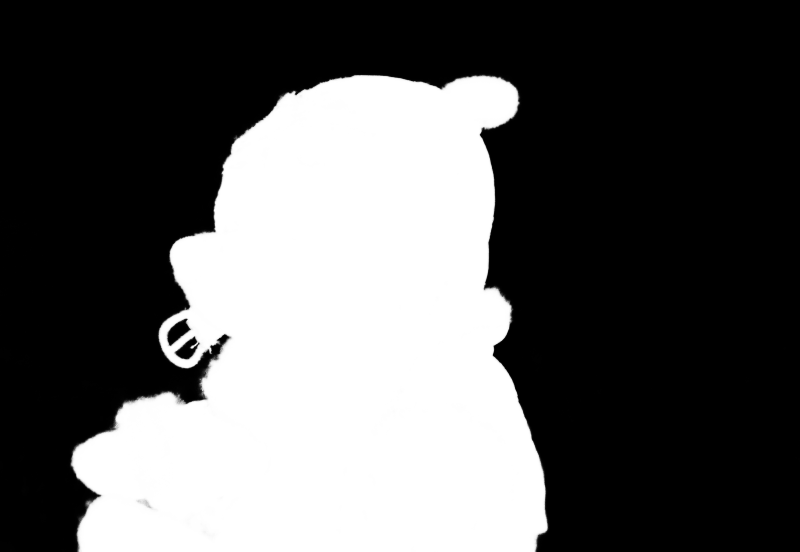}
  \end{subfigure}
  \hfill
  \begin{subfigure}{\sixwide\linewidth}
  \includegraphics[width=\textwidth]{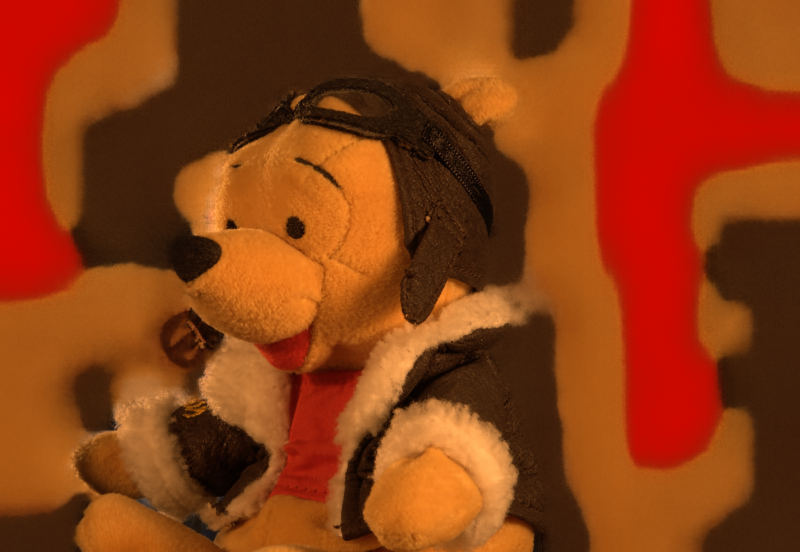}
  \end{subfigure}
  \hfill
  \begin{subfigure}{\sixwide\linewidth}
  \includegraphics[width=\textwidth]{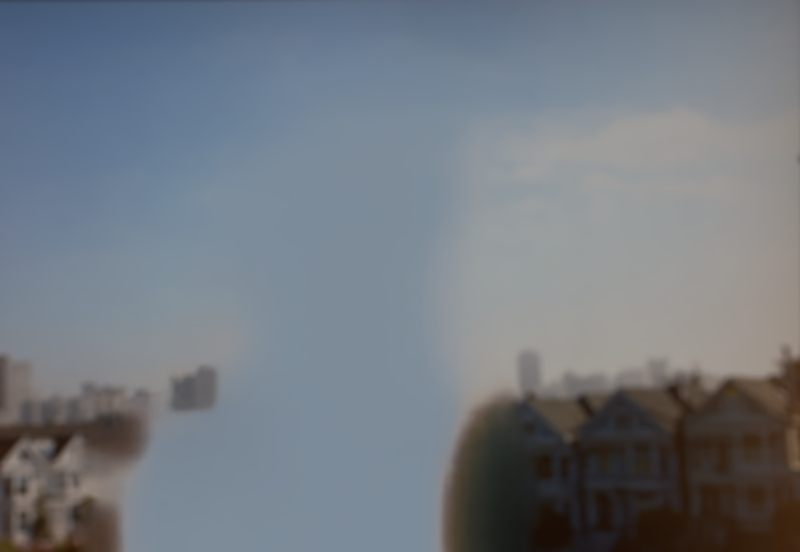}
  \end{subfigure}
  \caption{From left, image, trimap, GT alpha, estimated alpha, extrapolated foreground $\hat{F}$, extrapolated background $\hat{B}$.
     }
  \label{fig:showFandB}
\end{figure*}

\section{Background and Related Work}

The alpha-compositing equation is
\begin{equation}
I_i = \alpha F_i + (1-\alpha) B_i 
\label{eq:mattingeq}
\end{equation}
where $i \in \{r,g,b\}$, 
$I_i$ are red, green, and blue values at a pixel of the given image (given),
$F_i$ are the foreground object color (unknown),
$B_i$  are the background object color (unknown),
and $\alpha \in [0,1]$ is the (partially unknown) "alpha matte", representing the translucency or partial coverage of the foreground object at the pixel.

Typical images contain large areas that are relatively easily estimated to be solid "foreground" ($\alpha=1$) or "background" ($\alpha=0$),
as well as smaller regions where the alpha value is fractional. 
The matte extraction problem requires finding the these fractional alpha values.
The problem is underconstrained because there are only three known values $I_r,I_g,I_b$ but seven unknown values.
Hair is the prototypical challenge for matte extraction due to its irregular coverage and translucency,
though fractional alpha also generally appears along all object edges due to the partial coverage of a pixel by the foreground object.
It is also necessary to estimate the unknown foreground color in the fractional alpha regions, 
since without this the foreground cannot be composited over a different background.
Many methods assume that approximate demarcations of the solid foreground and background are provided,
e.g.~in the form of artist-provided "trimaps" \cite{alphamattingbenchmark_paper} or scribbles \cite{levin-naturalimagematting-06}.

While the research literature generally considers the problem of matte extraction from arbitrary natural backgrounds, 
in industry practice matte extraction from a greenscreen background is far from a solved problem and often requires an artist-curated
combination of techniques to obtain the required quality \cite{videomattingbenchmark_paper,artistcomment}.
LED walls remove the need for greenscreens in some cases, and are well suited for fast-paced television production,
however they also have drawbacks: computationally expensive physical or character simulations are not possible due to the need for real-time playback in a game engine,
bright backgrounds may introduce challenging light spill on the foreground physical objects \cite{fxguide-LEDwallpart1},
their cost is prohibitive for smaller studios, %
and the effects must be finished at the time of the principle shoot 
thus excluding the creative control and iterative improvement available in traditional post production.

Progress on matte extraction algorithms has been greatly facilitated by datasets that include 
ground truth (GT) mattes \cite{alphamattingbenchmark_paper,videomattingbenchmark_paper}.
The GT mattes have been obtained by different means including chroma keying from a greenscreen 
and photographing a representative toy object in front of a background image on a monitor (Fig.~\ref{fig:teaser} (a)).
Failure to closely approximate these GT mattes indicates a poor algorithm,
but an exact match may be unobtainable for several reasons:
1) the chroma key itself involves an imperfect algorithm \cite{videomattingbenchmark_paper},
2) light from multiple locations %
can physically scatter though (e.g.) %
translucent hair to arrive at a single pixel; 
this cannot be simulated in a purely 2D matte extraction process,
3) the image gamma or color space used in the benchmarks is not always evident and using a different gamma will produce small differences.

The underconstrained nature of matte extraction has been approached with a variety of methods \cite{wang-survey-07}.
One classic approach estimates the unknown alpha at a pixel based on similarity to the 
distributions of known foreground and background colors in solid regions \cite{Primatte-92,he-globalsampling-11}.
Another prominent principle finds the unknown alpha value by propagating from surrounding known values \cite{levin-naturalimagematting-06,aksoy-semanticsoft-18}. 
These approaches often require solving a system involving a generalized Laplacian formed from the pixel affinities in the unknown region,
which prevents real-time or interactive use.

Deep learning (DL) approaches are used in recent research e.g.~ \cite{semanticimagematting-sun-2021,BGMv2-uw-2021}. Much of this work focuses on 
providing alternate backgrounds for video meetings, and training databases consist of mostly forward-facing human heads. %
A state of the art method \cite{BGMv2-uw-2021} demonstrates high-quality matte extraction on HD-resolution images at 60fps. 
Many DL methods adopt a combination of techniques, e.g. separate networks for overall segmentation and for fractional alpha regions, or other hybrid approaches.

\subsection{Deep Image Prior}
The Deep Image Prior (DIP) \cite{DIP-ulyanov-2018} demonstrates that the architecture of an \emph{untrained} convolutional network 
provides a surprisingly good prior for tasks such as image inpainting and denoising. 
The key observation is that while a powerful DNN can fit arbitrary image structures such as noise, 
it is "easier" to fit natural image structures, as reflected in faster loss curve decay. 
Most experiments \cite{DIP-ulyanov-2018} optimize the weights of a U-net \cite{unet} that maps \emph{fixed} random noise to a \emph{single} output image. 
Uncorrupted features of the image are fit earlier in the optimization process, 
so \emph{early stopping} results in a corrected (denoised or inpainted) image.

The DIP has been successfully applied to other problems in image processing. Unsupervised coarse binary segmentation is demonstrated in \cite{gandelsman-doubledip-18}, based on a principle that it is easier (in terms of loss decay) 
to fit each component of a mixture of images with a separate DIP rather than using a single model. 
Concurrent with our work, \cite{xu-deepbackgroundmatting-22} formulate a DIP approach to \emph{background matting.} This problem scenario differs from our work in that it requires a clean plate, and hence is unsuitable for many VFX applications such as those that are filmed outdoors with a moving camera.

Our work also uses the DIP. We focus on high-quality estimation of fractional alpha mattes and, in contrast to previous work, undertake the challenging case where no clean plate is available, instead relying only on trimaps that can be 
easily created during post-production.

\section{Method}

Our method starts with a DIP network to reconstruct the target image. 
A second output head is added and tasked with inpainting the desired alpha in the trimap unknown region, constrained by the values in known regions.
The main idea is that the first output head forces the network representation preceding it to "understand the structure of the image," while the second output head makes use of that information in estimating the matte.

In addition to the alpha output we add two additional networks which simultaneously reconstruct the foreground and background. Similarly to the first output of the first network, these networks' outputs are constrained to match the target image but, unlike the first network, they are constrained only in their respective regions of the trimap. Similarly to the alpha output, they extrapolate those constraints to inpaint the unconstrained region (see Fig.~\ref{fig:showFandB}).

The latter three outputs -- $\hat{\alpha}, \hat{F}, \hat{B}$ -- are coupled via an additional constraint that they together satisfy 
\eqref{eq:mattingeq} (see \emph{Detailed Loss} below).
We put $\hat{F}$ and $\hat{B}$ in separate networks (as in \cite{gandelsman-doubledip-18})
rather than as additional outputs heads of the first network (as in \cite{semanticimagematting-sun-2021})
in order to allow the inpainting of the foreground to be independent of the background.
All three networks share the same generic U-net structure  \cite{DIP-ulyanov-2018}, except for an additional output head
in the first network. Our experiments use Adam \cite{kingma-adam} with a learning rate of 0.001.

\subsection{Detailed Loss}
The first term of our loss function is the reconstruction loss between the first network output and the target image:
\[
L_I = \frac{1}{|I|} \sum_{i \in I} \| \hat{I}_i - I_i \|^2
\]
The second loss term constrains $\hat{\alpha}$,
\[
L_{\alpha} = \frac{1}{|C|} \sum_{i \in C} \| \hat{\alpha}_i - T_i \|^2
\]
where $T$ is the trimap and $C = F \cup B$ is the \textit{constrained} region of the trimap.

The reconstruction losses for the foreground and background outputs are defined similarly to $L_I$, but constrained only in their respective regions of the trimap:
\[
L_F = \frac{1}{|F|} \sum_{i \in F} \| \hat{F}_i - I_i \|^2
\qquad
L_B = \frac{1}{|B|} \sum_{i \in B} \| \hat{B}_i - I_i \|^2
\]

The three networks' outputs are coupled via the alpha-compositing equation as follows:
\[
L_c = \frac{1}{|U|} \sum_{i \in U} \| I_i - (\hat{\alpha}_i \cdot \hat{F}_i + (1-\hat{\alpha}_i) \cdot \hat{B}_i) \|^2
\]
where $U = I - C$ is the \textit{unconstrained} region of the trimap.

Finally, we include an exclusion loss similar to \cite{semanticimagematting-sun-2021} to prevent the structure of the foreground from leaking into the background and vice-versa:
\[
L_e = \frac{1}{|U|} \sum_{i \in U} \| \nabla \hat{F}_i \|_1 \| \nabla \hat{B}_i \|_1 + \| \nabla \hat{\alpha}_i \|_1 \| \nabla \hat{B}_i \|_1
\]

The total loss is the sum of the above six components:
\[
L = L_I + L_{\alpha} + L_F + L_B + L_m + L_e
\]
Unlike other DIP-based techniques, our algorithm does not require early stopping since the goal is to exactly fit both the image and the trimap constraints.

\begin{figure*}
\centering
\begin{minipage}{.55\textwidth}
\newcommand{\figresultswid}{0.227}  %
\addtolength{\tabcolsep}{-3pt}
 \centering
 \begin{tabular}{cccc}
  \begin{subfigure}{\figresultswid\linewidth}
  \includegraphics[width=\textwidth]{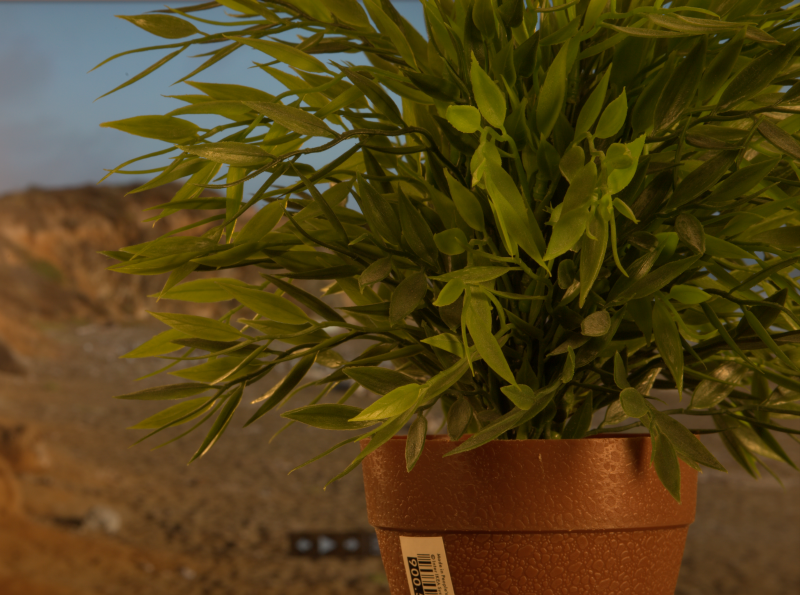}
  \end{subfigure}
    &
  \begin{subfigure}{\figresultswid\linewidth}
  \includegraphics[width=\textwidth]{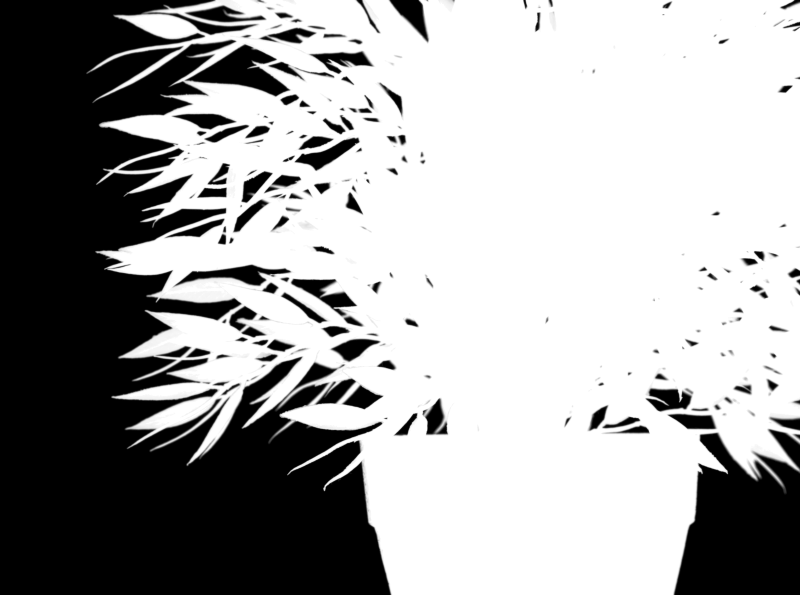}
  \end{subfigure}
    &   %
    \begin{subfigure}{\figresultswid\linewidth}
  \includegraphics[width=\textwidth]{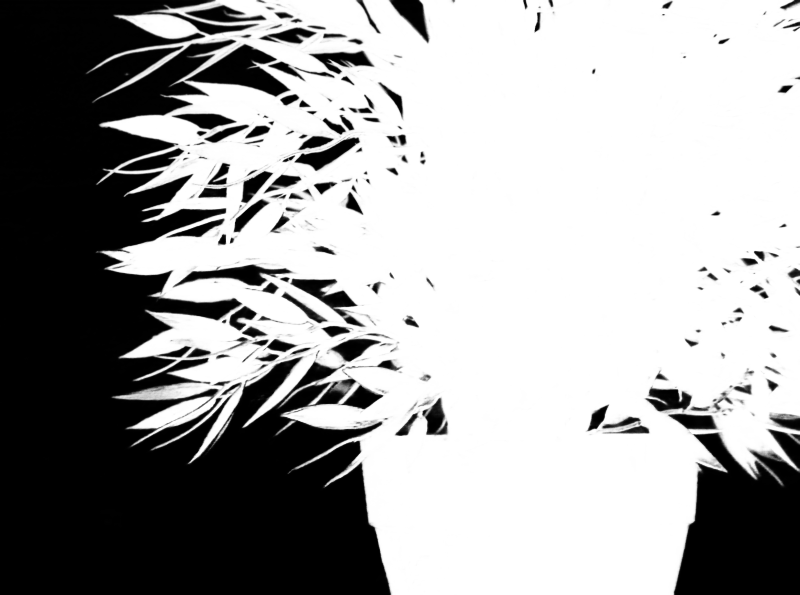}
  \end{subfigure}
    &
  \begin{subfigure}{\figresultswid\linewidth}
  \includegraphics[width=\textwidth]{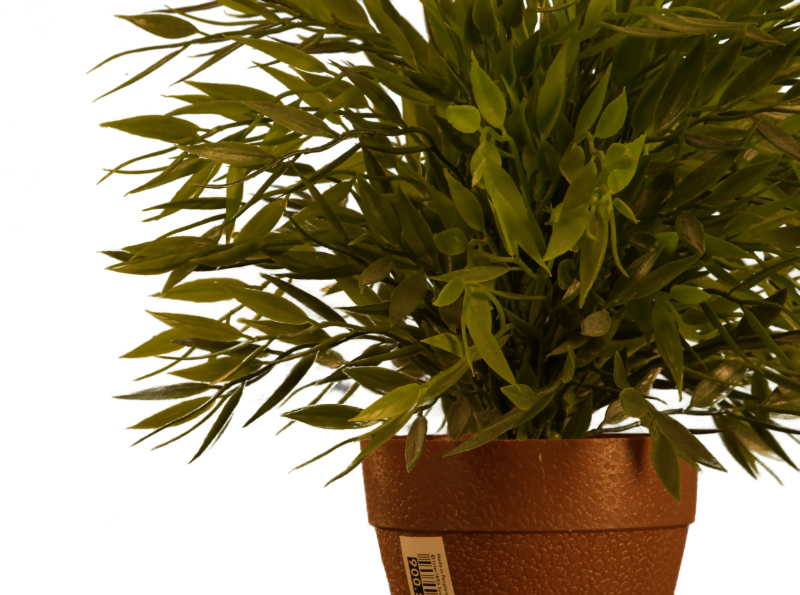}
  \end{subfigure}
\\[19pt]
  \begin{subfigure}{\figresultswid\linewidth}
  \includegraphics[width=\textwidth]{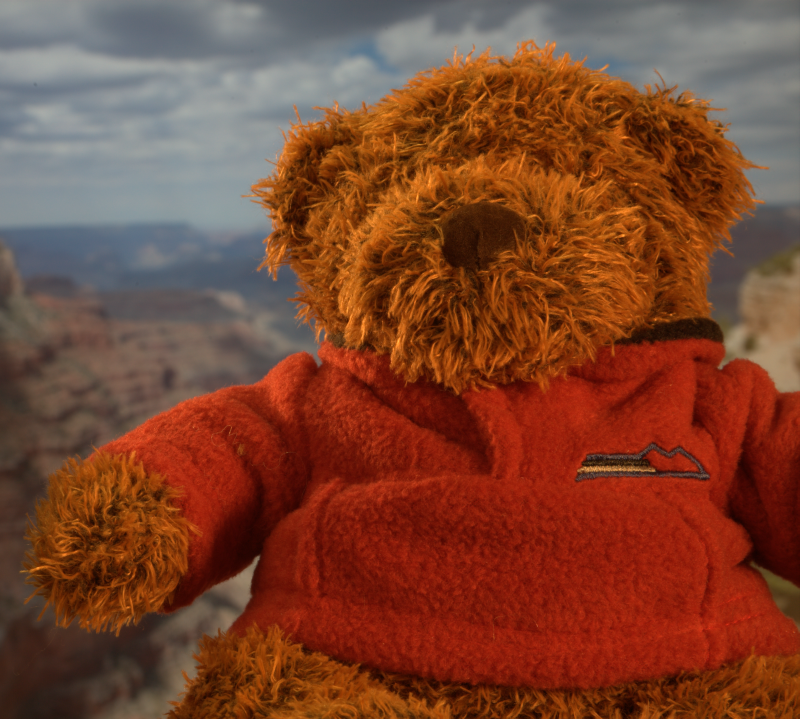}
  \end{subfigure}
     & 
  \begin{subfigure}{\figresultswid\linewidth}
  \includegraphics[width=\textwidth]{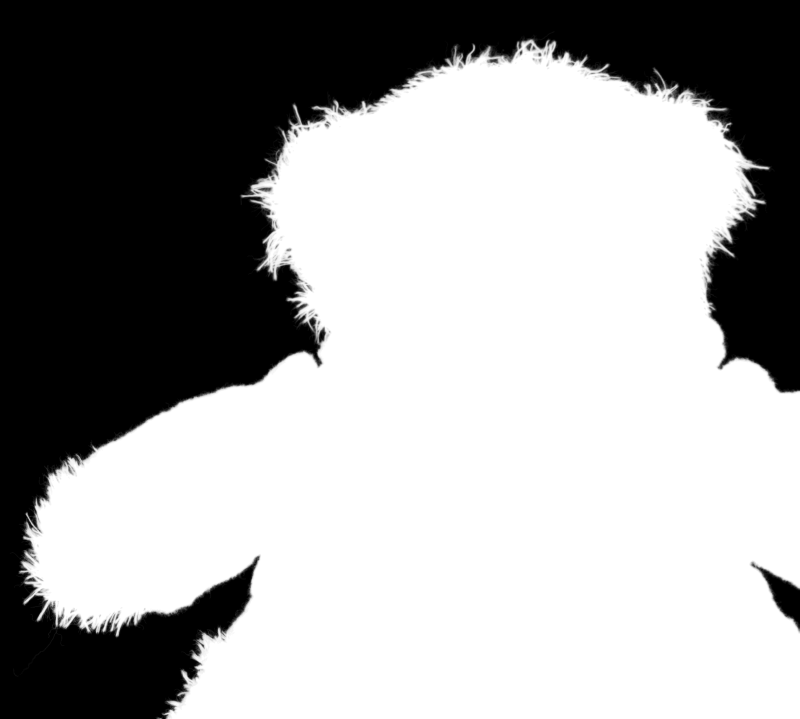}
  \end{subfigure}
     &
  \begin{subfigure}{\figresultswid\linewidth}
  \includegraphics[width=\textwidth]{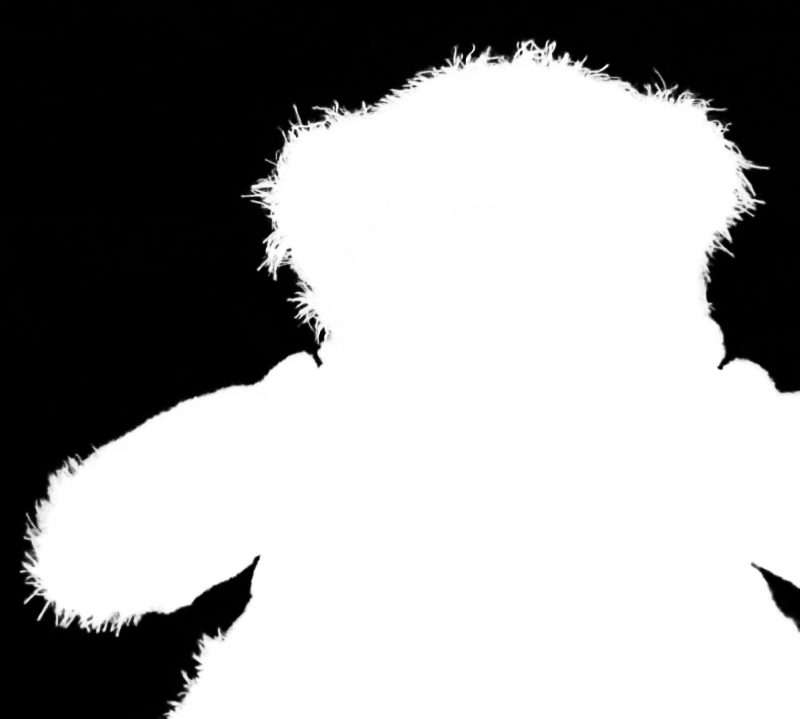}
  \end{subfigure}
     &
  \begin{subfigure}{\figresultswid\linewidth}
  \includegraphics[width=\textwidth]{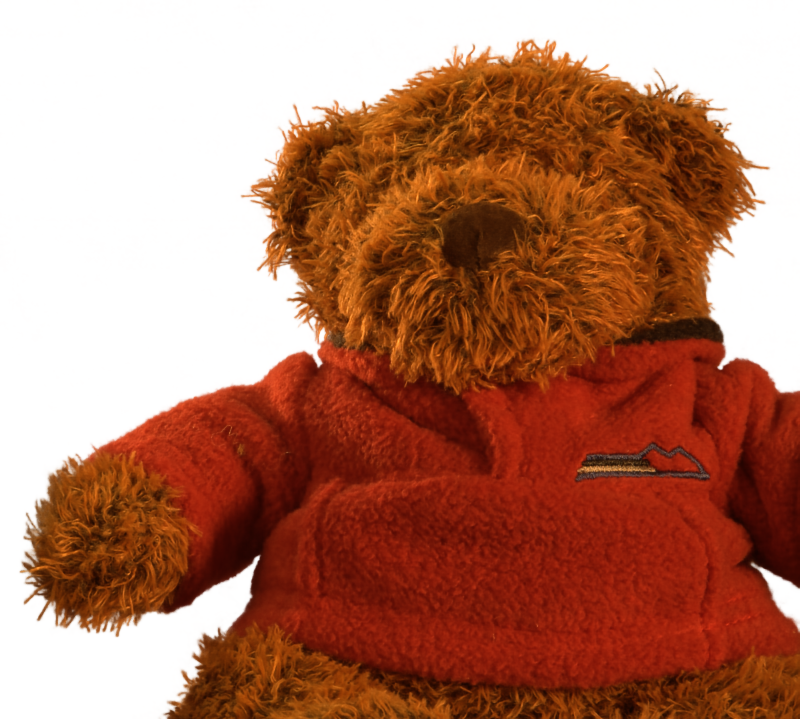}
  \end{subfigure}
     \\
 \end{tabular}
  \caption{
         From left: image, ground truth alpha map, estimated alpha map, and composite over white. Please enlarge to see details.
     }
  \label{fig:results1}
\addtolength{\tabcolsep}{3pt}

\end{minipage}%
\hfill
\begin{minipage}{.4\textwidth}
\input{FIGfailurecase}
\end{minipage}
\end{figure*}

\subsection{Temporal continuity}
In our experiments temporal continuity was obtained by warm-starting the optimization 
for frames other than the first
with the final weight values of the previous frame, and stopping with a loss threshold rather than a fixed number of iterations (see video).
This simple strategy produces reasonable results even on the relatively difficult case of hair.
It also reduces the compute time by roughly an order of magnitude.

\subsection{Why does it work?}
The deep image prior empirically demonstrates that standard convolutional networks can act as a good low-level prior for image reconstruction tasks,
but why can it be adapted to invent plausible alpha mattes?
Our intuition is as follows:
The DIP provides a somewhat deep and hierarchical encoding of the particular image.
The features in this encoding will primarily span valid image structures, with noise lying outside this space. 
In our work these features are re-combined to produce the matte.
This works because the alpha matte contains structures that relate to those in the foreground image. 
In this respect our approach resembles a "deep" version of the guided image filter \cite{he-guidedimagefilter-10}.

\section{Results and Failure Cases}

We use images, trimaps, and GT alpha from the dataset  \cite{alphamattingbenchmark_paper} to allow comparison to the ground truth.
Fig.~\ref{fig:teaser} shows a moderately challenging case with extensive hair, some of which is similar to the background color.
Fig.~\ref{fig:results1} shows several additional examples.
While the ground truth and estimated alpha maps appear identical at first glance, small differences can be seen upon enlargement. These may be due to imperfections in our algorithm, but also may reflect differences in color spaces 
between our algorithm and the ground-truth estimation process. %

Fig.~\ref{fig:showFandB} shows the extrapolated foreground $\hat{F}$ and background $\hat{B}$ for a particular image. %
Note that the inpainted colors need to be correct only in the
unconstrained region surrounding the object border
in order to allow compositing over a new background, as is the case in this figure.
The unrealistic extrapolated $\hat{F}$ over the pure background regions is ignored.

\subsection{Failure cases}
While objects with "holes" sometimes %
yield
good mattes (e.g.~the plant in Fig.~\ref{fig:results1}), they are also a failure case (Fig.~\ref{fig:screenfailure}).
In some cases the trimap can be adjusted to highlight the missing holes,
though
this would be laborious in cases such as the cup %
in Fig.~\ref{fig:screenfailure}.

\section{Limitations and Conclusion}

We have introduced a matte extraction approach using the deep image prior.
The algorithm is simple, requiring only a few tens of lines of code modification to an existing U-net.
Our approach is training-free and is thus particularly suitable for %
the diverse, few-of-a-kind subjects in entertainment video production.
It also may be of intrinsic theoretical interest in terms of the nature and solution of the matte extraction problem.
A further potential use would be to produce ground-truth mattes to be used for DL training. As is the case with many matting algorithms, it assumes coarse guidance in the form of a trimap or similar constraints.
This can be created by the artist using readily available semi-automatic tools. %

Computational cost is the major limitation of the method,
in common with classic methods \cite{levin-naturalimagematting-06}.
Compute times for the examples shown in the paper are measured in minutes (but not hours) on a single previous generation Nvidia Volta GPU. 
This restricts the use of our algorithm to high-quality offline applications where extensive non-real-time computation is the norm, 
primarily movies and videos. On the other hand, the computation can take advantage of support for multiple GPUs provided in deep learning frameworks, and intermediate results can be visualized. 

Our method can produce temporally consistent matte extractions from video by warm-starting the optimization from the previous frame (see accompanying video), however in our experience this requires that the trimaps have smooth motion from frame-to-frame. A topic for future work is to consider recurrent or other 
network 
architectures that might make the trimap choice more forgiving.
This paper has focused on introducing the DIP matting algorithm. There was relatively little architecture and parameter exploration, and further improvements may be possible.

\section*{Acknowledgments}
G.G.~Heitmann, Peter Hillman, and Kathleen Beeler gave helpful insights and feedback.

\bibliographystyle{ACM-Reference-Format}
\bibliography{REFERENCES}

\end{document}